\documentclass[doublecol]{epl2} 
\usepackage{graphicx}
\usepackage{epsfig}
\usepackage{subfigure}
\usepackage{amsmath,amsfonts,amssymb,graphicx,times}
\usepackage{algorithm}
\usepackage{algpseudocode}
\usepackage{diagbox}
\usepackage{float}
\usepackage{booktabs}
\usepackage[colorlinks=true,linkcolor=black,citecolor=green,urlcolor=green]{hyperref}
\title{The State-Action-Reward-State-Action Algorithm in Spatial Prisoner's Dilemma Game}
\shorttitle{The SARSA Algorithm in Spatial Prisoner's Dilemma Game}

\author{Lanyu Yang\inst{1} \and Dongchun Jiang\inst{1} \and Fuqiang Guo\inst{1} \and Mingjian Fu\thanks{Corresponding author: \email{sinceway@fzu.edu.cn}}\inst{1,2}}
\shortauthor{L. Yang \etal}

\institute{
  \inst{1} College of Computer and Data Science,
  Fuzhou University, Fuzhou, 350116, China\\
  \inst{2} Fujian Key Laboratory of Data Science and Statistics,
  Minnan Normal University, Zhangzhou 363000, China
}
\pacs{02.50.Le}{Decision theory and game theory}
\pacs{87.23.Kg}{Dynamics of evolution}
\pacs{87.23.Ge}{Dynamics of social systems}

\abstract{Cooperative behavior is prevalent in both human society and nature. Understanding the emergence and maintenance of cooperation among self-interested individuals remains a significant challenge in evolutionary biology and social sciences. Reinforcement learning (RL) provides a suitable framework for studying evolutionary game theory as it can adapt to environmental changes and maximize expected benefits. In this study, we employ the State-Action-Reward-State-Action (SARSA) algorithm as the decision-making mechanism for individuals in evolutionary game theory. Initially, we apply SARSA to imitation learning, where agents select neighbors to imitate based on rewards. This approach allows us to observe behavioral changes in agents without independent decision-making abilities. Subsequently, SARSA is utilized for primary agents to independently choose cooperation or betrayal with their neighbors. We evaluate the impact of SARSA on cooperation rates by analyzing variations in rewards and the distribution of cooperators and defectors within the network.}

\begin{document}

\maketitle
Keywords: reinforcement learning, SARSA, evolutionary game theory, prisoner’s dilemma, cooperation

\section{Introduction}\label{Introduction}
According to Darwin's theory of natural selection, over-reproduction is a common tendency among most organisms. But the resources required for survival are limited, resulting in intense inter-individual competition~\cite{darwin1}. Cooperative behavior can provide significant benefits to a population; however, it may reduce the competitive advantage of individuals. As a result, understanding how cooperative behavior arises and is maintained among self-interested individuals has become a critical area of exploration in both theoretical and empirical research~\cite{fields1,fields2,fields3,fields4}.

Evolutionary game theory provides a theoretical framework for understanding the emergence and maintenance of cooperative behavior among self-interested individuals~\cite{Evolution1}. In this framework, initial individuals are assumed to be irrational and must to try and adjust their strategies~\cite{Evolution2, Evolution3, Equilibrium}. The qualified dilemma model takes participants as the research object and gives them a fixed strategy set. After specifying the initialization conditions, all participants interact with their opponents in each epoch of the game to obtain corresponding benefits and update their strategies according to the learning rules. Repeating the above process, all individuals in the game system eventually reach a dynamic evolutionary stable equilibrium point~\cite{equilibrium1, equilibrium2, equilibrium3}. The study of the existence of cooperative behavior in fields such as ecology, economy, and human society has significant practical importance. The development of evolutionary game theory based on complex networks provides a clear theoretical framework for solving practical problems, and its application in various fields has attracted increasing attention in recent years~\cite{Gapplication,equilibrium2,hammerstein1}.

In recent years, there has been a significant advancement in artificial intelligence technology. Reinforcement learning(RL), a subset of artificial intelligence, operates on the principle of learning through trial and error when interacting with the environment~\cite{fundamental1, fundamental2, fundamental3, fundamental4, fundamental5}. To maximize the desired reward, agents continuously "try and error" their learning processes. With the development of RL technology, RL algorithms have achieved a general intelligence that can solve complex problems to a certain extent. In game genres like Go, it can even compete at a human level. When playing chess against itself, AlphaGo Zero employs the RL algorithm, which enables the system to develop from an initial blank state where it doesn't understand the rules to a master who eventually defeats AlphaGo~\cite{AlphaGo}. RL has proven to be effective in solving game problems in the real world~\cite{solveProblems}. One of the most popular research areas is the application of RL algorithms to the evolutionary dynamics of cooperative behavior in game theory~\cite{direction}. The main objective of this paper is to apply the State-Action-Reward-State-Action (SARSA) algorithm to evolutionary game theory and investigate the impact of the SARSA algorithm on network cooperation rates. SARSA is a table-based RL algorithm that is widely used for solving reinforcement learning problems.

\section{Models and Methods}\label{Models and Methods}
The model can be described as follows: agents are placed on an $L \times L $  square lattice with periodic boundaries. In each epoch, participants choose to cooperation or betrayal simultaneously. In this paper, we use the prisoner's dilemma, a theory related to the dilemma formulated by Merrill Flood and Melvin Dresher of RAND Corporation in 1950~\cite{Pdilemma}. For agent i, its reward is defined as follows: 

\begin{center}
	\begin{tabular}{ c | c | c }
		\toprule[1.5pt]
		\diagbox{Agent $i$}{Neighbor agent}  & Cooperate & Betray \\
		\midrule[1pt]
		Cooperate & $(R, R)$ & $(S, T)$ \\
		\midrule[1pt]
		Betray & $(T, S)$ & $(P, P)$ \\
		\bottomrule[1.5pt]
	\end{tabular}
\end{center}

The values in the reward matrix satisfy the condition $T > R > P > S$. In the prisoner's dilemma, when both parties do not know the other's choice, the best strategy for each suspect is to betray the other and plead guilty to minimize their sentence and maximize their reward. The Nash equilibrium state of this game is when both suspects confess to the crime, meaning that no individual can increase their reward by changing their strategy. Therefore, betraying each other is the best choice for the individual. However, if both suspects choose to remain silent, this will be the worst outcome in terms of collective interests. The prisoner's dilemma describes the conflict between individual and collective interests, that is, the social dilemma. In the experiment, the reward matrix is as follow:

\begin{equation}
	\left[ \begin{matrix}
		R & S\\
		T & P
	\end{matrix} \right] = 
	\left[\begin{matrix}
		1 & -D_r\\
		1 + D_g & 0
	\end{matrix} \right]
\end{equation}

Here $ 0 \leq D_g \leq 1 $ and $ 0 \leq D_r \leq 1 $.

The algorithm used in this study is the SARSA algorithm, which is a RL algorithm. SARSA was first proposed by G.A. Rummery and M. Niranjan in 1994~\cite{SARSA1}, and the concept was formally introduced by R.S. Sutton in 1996~\cite{SARSA2}. By using the SARSA algorithm, agents can improve their ability to select the best action based on the action sequence and the corresponding reward value.

In the SARSA algorithm, the interaction process between the agent and the environment can be regarded as a Markov decision process with finite state and discrete-time:$ M = <S, A, R, P> $. Here, $ S = \{ s_1, s_2, …, s_n \} $ represents the state set, $ A = \{ a_1, a_2, …, a_n \} $ represents the action set, $ R : S \times A \rightarrow R$ represents the reward function and $P:S \times A \rightarrow [0,1]$ represents the state transition function. In epoch $t$, the agent observes the current state $s_t$, selects the action $a_t$ to interact with the environment, and receives a reward value $r_t$. The agent learns the experience value according to the reward value and observes the state $s_{t+1}$. Through this process, the agent gradually becomes more intelligent.

The fundamental concept of SARSA is similar to Q-learning~\cite{qlearning}in that it also utilizes the evaluation of the state-action pair's value, denoted as $Q (s_t, a_t)$, to make decisions. Here, Q refers to the estimated reward value obtained when the agent selects action $a_t$  through the strategy function at state $s_t$. The primary difference between SARSA and Q-learning is that SARSA is on-policy. In other words, during every learning iteration, it updates its Q-value based on the actual actions $a_t$ it takes.

\begin{equation}
\begin{aligned}
	Q(s_t,a_t )=&Q(s_t,a_t ) + \alpha \{ r_t + \\
	& \gamma Q(s_{t+1},a_{t+1} )-Q(s_t,a_t )\} \label{sarsa-update}
\end{aligned}
\end{equation}
\begin{equation}
\begin{aligned}
	Q(s_t,a_t )=&Q(s_t,a_t ) + \alpha \{r_t + \\
	& \gamma \max_{a'} Q(s_{t+1},a' )-Q(s_t,a_t )\}
\end{aligned}
\end{equation}

Equation (2) and (3) correspond to the Q-value update methods for SARSA and Q-learning, respectively. In order to balance the exploration and exploitation, both the SARSA and Q-learning algorithms incorporate the $\varepsilon$-greedy stategy during the learning process. Here, $\alpha$ represents the learning rate, which constrains the amount of learning each iteration, and $\gamma$ represent discount rate, which constrains the influence of past experiences. For Q-learning, each time the Q-value updates, it selects the item with the largest estimated value of state $s_{t+1}$ reward in the experience to update. As a result, Q-learning agents tend to overestimate the reward value, and thus their evaluation of the reward value is often high. By contrast, the SARSA algorithm may be more suitable for the game environments, as it updates Q-values based on the agents' selected actions which provides a more human-like approach to experience acquisition.

In the paper, we classify game participants as traditional agents and SARSA agents. To provide a point of comparison, traditional agents use only the Fermi update rules, while SARSA agents incorporate this formula for final policy checks. The formula is presents as follows:

\begin{equation} 
	\label{Fermi-update}
	W_{S_x \leftarrow S_y} = \frac{1}{ 1 + e^{\frac{{\pi}_x-{\pi}_y}{K}} }
\end{equation}

Here $x$ and $y$ represent the central individual and one of its neighbors in the network. $S_x$ and $S_y$ represent the strategies of $x$ and $y$. ${\pi}_x$ and ${\pi}_y$ are the rewards of $x$ and $y$. $K$ stands for the noise factor. When $K \rightarrow 0$, it means that the individual is completely rational. As long as $r_x > r_y$, individual will directly learn $y$'s strategy. When $K \rightarrow \infty$, individuals will learn the strategy of $y$ with a probability of 1/2. Here the noise $K$ is set to be 0.1.

\section{Simulation and Analysis}\label{Simulation and Analysis}

\begin{figure*}[htbp]
	\centering{ \includegraphics[scale=0.6]  {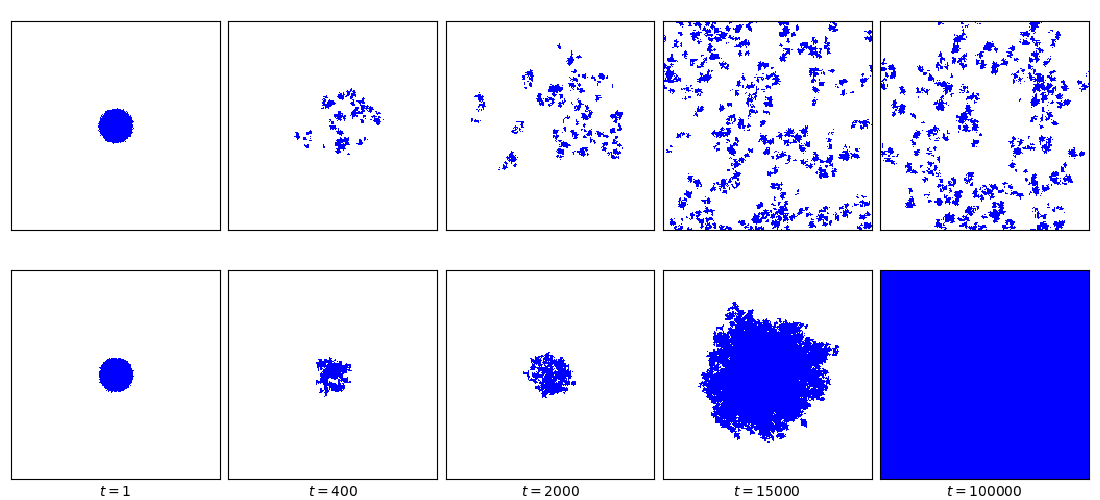 }}
	\caption{ \label{bantu1} Spatial distribution of cooperators (blue) and defectors (white) at different time steps when the Prisoner's dilemma parameter $D_r=0$, $D_g=0.02$. The first row is the evolution diagram of traditional individuals, and the second row shows the evolution diagram for agents trained using the SARSA algorithm.}
\end{figure*}

The experiment was designed based on the methodology described in Reference~\cite{method1}, wherein agents were programmed to follow the decisions of their higher-earning neighbors. To learn the optimal strategy, each agent in the square lattice utilized the SARSA algorithm to select a neighboring agent and incorporate its strategy. This approach aimed to simulate the decision-making behavior observed in society.
The specific process is described as follows: In each epoch, agents first observe the cooperation rate between themselves and their neighbors to determine the current state, denoted as S. Subsequently, using the SARSA algorithm, each agent selects a neighboring agent as a learning object from the adjacent area. Once the learning object is determined, the agent decides whether to adopt its strategy based on the Fermi update rule, which was utilized in the traditional method described earlier.

In the following section, all the numerical results, which are obtained from the Monte Carlo (MC) simulation, are carried on a $L \times L$($L=200$) square lattices with periodic boundary. The primary parameters of the algorithm were set as follows: $\alpha=0.3$, $\gamma=0.9$, and $\epsilon=0.02$. 

The spatial distribution of cooperators and defectors at different time steps depicted in Figure \ref{bantu1} illustrates that cooperators trained with the SARSA algorithm tend to form clusters, making it difficult for defectors to coexist with them. Initially, all cooperators were located at the center. Although defectors diluted cooperative behavior along the border, it did not compromise the structure of the cooperative cluster. Thus, even when defectors exploited cooperators, they remained connected. Over time, the cooperative cluster gradually expanded and ultimately dominated the network, leading to the complete elimination of defectors.

Thus, it can be inferred that the SARSA mechanism leads to the information of tightly-knit cooperative clusters, even when starting with the same initial cooperation rate. As the clusters expand over time, they outcompete defectors and ultimately dominate the network. 

\begin{figure*}[htbp]
	\centering{\includegraphics[width=14cm]  {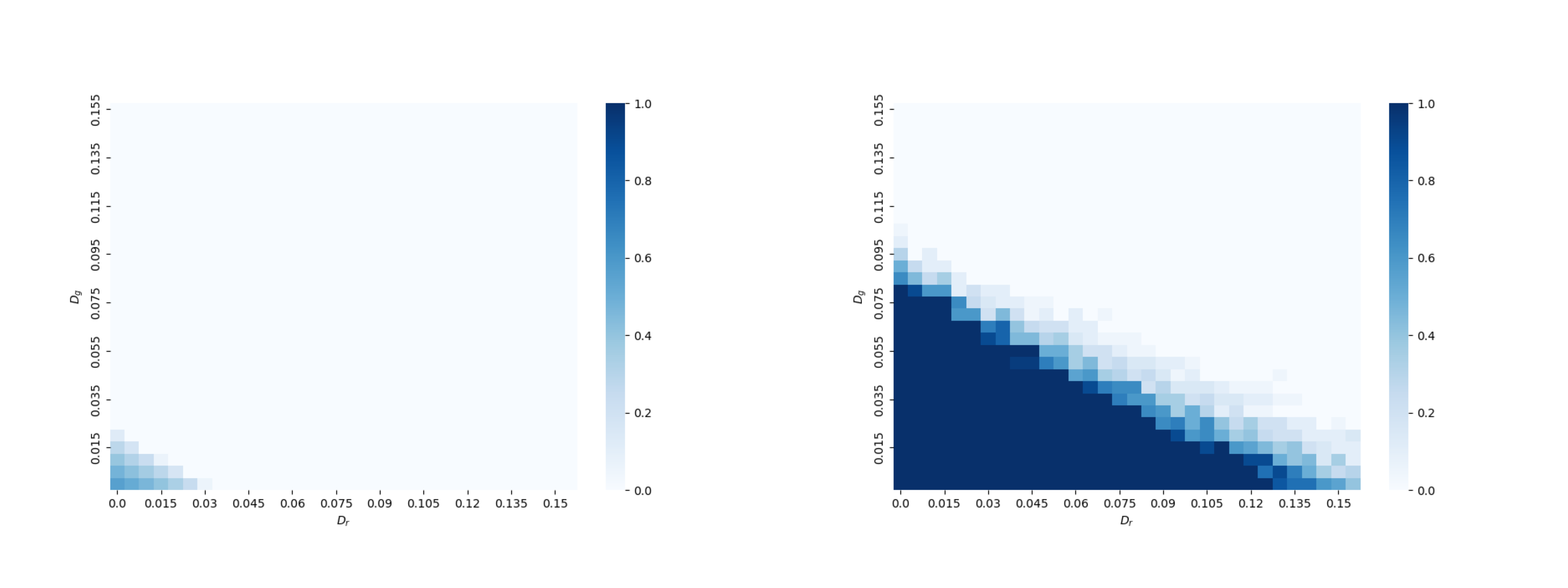}}
	\caption{\label{finally_DrDg}Heat map of cooperation rate under different $D_r$ and $D_g$.The graph on the left is traditional, and the other use the SARSA algorithm.}
\end{figure*}

Fig.\ref{finally_DrDg} shows the final cooperation rate of two types of agents under different $D_g$ and $D_r$ parameters. 

\begin{algorithm}[H]
	\caption{: The Evolutionary Algorithm of Choosing Target Object}
	\label{alg:1}
	\begin{algorithmic}[1]
		\Require Initialize $\alpha$,$\gamma$,$\epsilon$, the Q-table and  strategy of all agents, random actions $a$; 
		\For{ epoch $t$ from $1$ to $T$ }:
		\For{agent $i$ }:
		\State $s^i_t \leftarrow s^i_{t+1}, a^i_{t} \leftarrow a^i_{t+1}, r^i_t \leftarrow r^i_{t+1}$;
		\State According to the cooperation rate of agent $i$ and its neighbors, find the current state $s^i_{t+1}$ of agent $i$; 
		\If{$random(0, 1)  > \epsilon $ }:
		\State $a^i_{t+1}$ is the one that maximizes $Q$, i.e. $a^i_{temp} \leftarrow \mathop{\arg \max}\limits_{a^i} Q(s^i_{t+1}, a^i)$.
		\Else:
		\State $a^i_{temp}$ will be random (one of neighbors).
		\EndIf
		\If{ $W_{S_x \leftarrow S_y}  \geq random(0, 1)$ }
		\State $a^i_{t+1} \leftarrow a^i_{temp}$
		\Else:
		\State  $a^i_{t+1} \leftarrow a^i_t$
		\EndIf
		\EndFor
		\State Then agent $i$ will use the strategy of neighbor $a^i_{t+1}$ to interact with its four neighbors in pairs to obtain a cumulative reward $r^i_{t+1}$;
		\State Agent $i$ updates the Q-table according to the information it knows, i.e. Eq.(\ref{sarsa-update});
		\EndFor
	\end{algorithmic}
\end{algorithm}	
Here we run the program 20 times using each parameter, i.e., the step size of $D_g$ and $D_r$ set as 0.01, and then take the average of the final cooperation rate. If the cooperation rate eventually does not reach 0 or 100\% but stays around a stable value, we will take the average of the last thousand rounds of cooperation after 500,000 epochs. From the figure, we can see that the SARSA mechanism will improve the survival probability of the cooperator in higher distress conditions. During the experiment, we found that if the cluster can survive successfully, the cooperator will eventually fail to survive.

\begin{figure}[htbp]
	\centering{\includegraphics[width=7cm]  {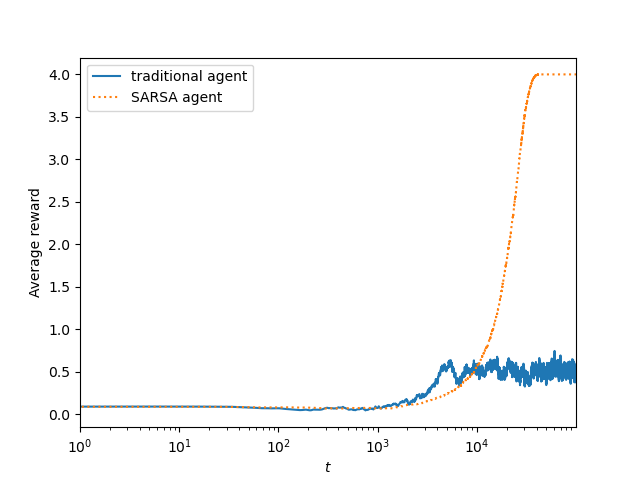}}
	\caption{\label{total_average}Average reward of traditional and SARSA.}
\end{figure}

\begin{figure*}[htbp]
	\centering{\includegraphics[width=14cm]  {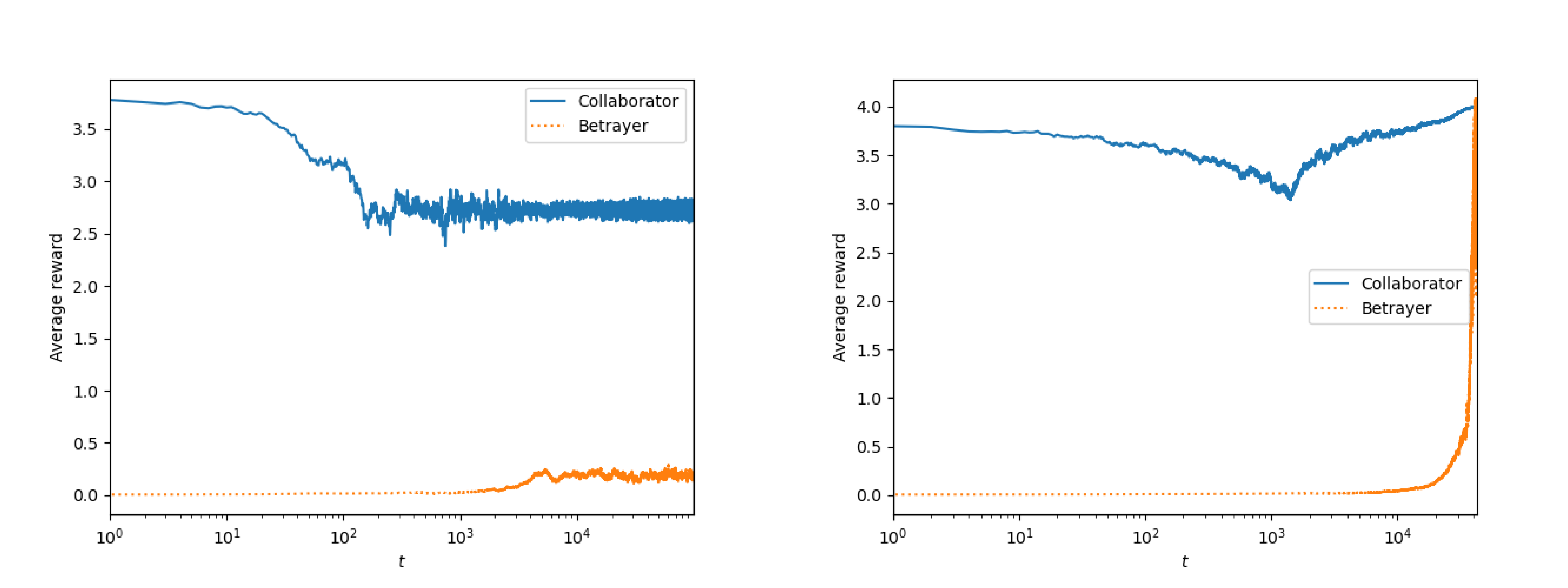}}
	\caption{\label{reward_compare}Average reward of cooperators and betrayers. The graph on the left is traditional, and the other use the SARSA algorithm.}
\end{figure*}

To explore why the traitor cannot conquer the cooperators on the border, we study the changes in rewards over time. Fig.\ref{total_average} is a comparison of the benefits of traditional agents and SARSA agents. We can find that over time, the average reward of SARSA agents is getting higher and higher, and in the end, it is much greater than that of traditional agents. Fig.\ref{reward_compare} shows the benefits of cooperators and betrayers between original agents and SARSA agents. Finally, they are stable near a value. However, the reward of agents using the SARSA mechanism will increase whether they are cooperators or betrayers. The betrayer reward is reduced to 0 because the betrayers disappear there.

\begin{algorithm}[H]
	\caption{: The Evolutionary Algorithm of Making the Strategy}
	\label{alg:2}
	\begin{algorithmic}[1]
		\Require Initialize  $\alpha$,$\gamma$,$\epsilon$, the Q-table of all agents, initialize action $a$.
		\For{ epoch $t$ from $1$ to $T$ }:
		\For{agent $i$}:
		\State $s^i_t \leftarrow s^i_{t+1}, a^i_{t} \leftarrow a^i_{t+1}, r^i_t \leftarrow r^i_{t+1}$;
		\State According to the cooperation rate of agent $i$ and its neighbors, find the current state $s^i_{t+1}$ of agent $i$; 
		\If{$random(0, 1)  > \epsilon $ }:
		\State $a^i_{t+1}$ is the one that maximizes $Q$, i.e. $a^i_{t+1} \leftarrow \mathop{\arg \max}\limits_{a^i} Q(s^i_{t+1}, a^i)$.
		\Else:
		\State $a^i_{t+1}$ will be random (cooperate or betray).
		\EndIf
		\EndFor
		\State Then agent $i$ will use the action $a^i_{t+1}$ to interact with its four neighbors in pairs to obtain a cumulative reward $r^i_{t+1}$;
		\State Agent $i$ updates the Q-table according to the information it knows, i.e. Eq.(\ref{sarsa-update});
		\EndFor
	\end{algorithmic}
\end{algorithm}	

In addition to the above work, we let agents make decisions based on their historical returns. In this part, agents' actions will be directly the strategy, i.e., betrayal or cooperation. There will exist on the network $\rho$ proportional agents, which can make decisions based on the SARSA algorithm. And define the other agents as traditional individuals(they apply algorithm 1 to get strategies). It emulates people's decision-making based on their own experience.

We can describe the specific process as follows. In epoch $t$, agents first need to determine their current state $s_t$ according to the cooperation rate between themselves and their neighbors. Then, they select action $a_t$ according to the SARSA algorithm. The action here refers to the cooperation or betrayal strategy adopted in the next epoch.

The parameter setting of the algorithm is $\alpha =0.3$, $\gamma =0.9$, and $\epsilon =0.02$. For the SARSA agent, the algorithm flow is as follows:

The first and second rows and the third and fourth rows of subgraphs are the evolution of the network under the same initial distribution conditions. We can see from the figure that under higher $D_g$ and $D_r$, the individuals of SARSA agents can survive better. And traditional agents attached to SARSA agents also tend to become cooperators. We can see that the SARSA mechanism can promote traditional agents to become cooperators.

Since the increase of $D_g$ and $D_r$ will increase the difficulty of cooperators' survival, we unify them as dilemma strength ($DS$), i.e., $DS = D_g = D_r$. Then we plotted the change curve of the final cooperation rate with the proportion of SARSA agents under different DS conditions.

\begin{figure*}[htbp]
	\centering{\includegraphics[width=14cm]  {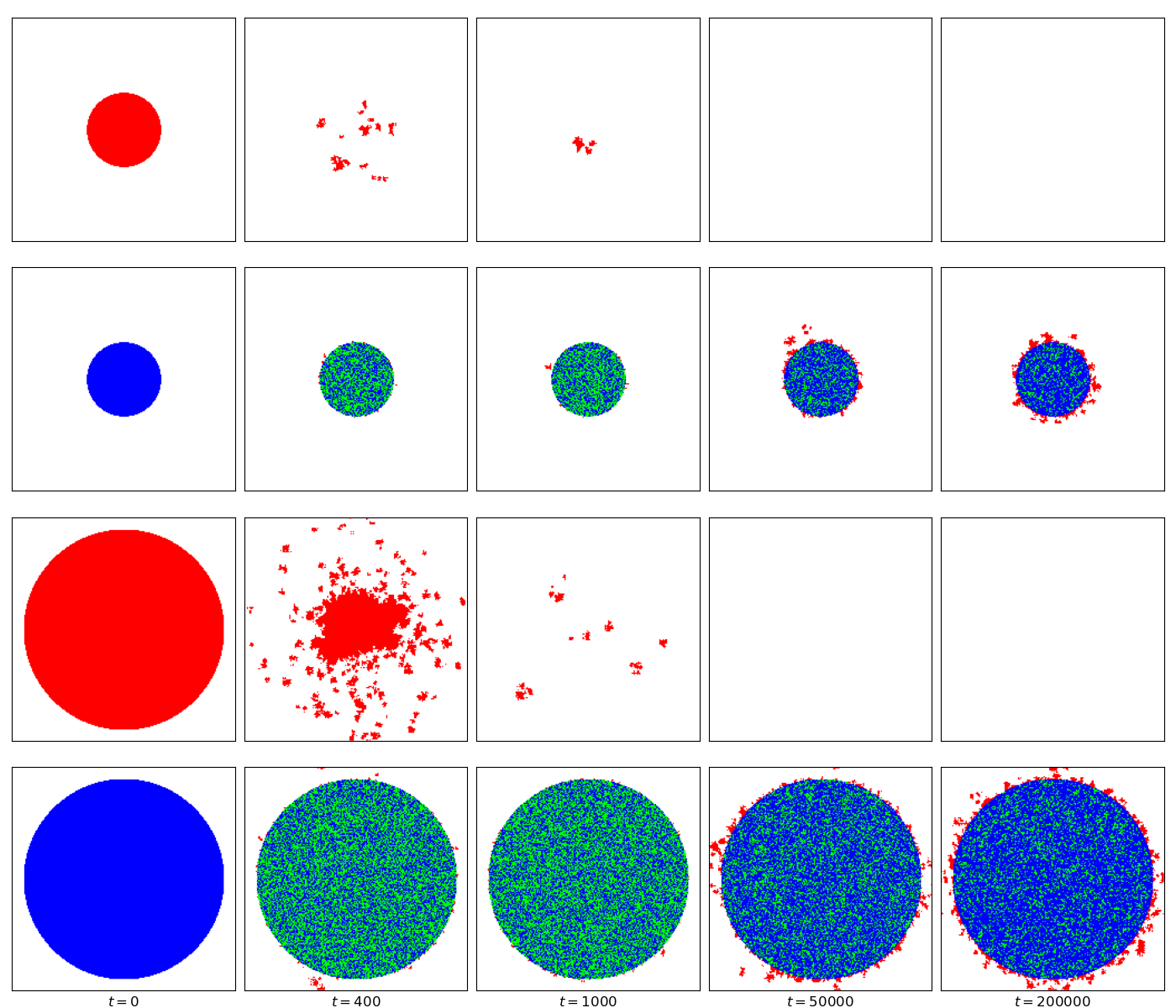}}
	\caption{\label{bantu2}Game evolution spot graph. White represents this individual as a betrayer but not the SARSA agent. Red represents this individual as a cooperator but not the SARSA agent. Blue represents this individual as a cooperator and the SARSA agent. Green represents this individual as a betrayer and the SARSA agent.}
\end{figure*}

\begin{figure*}[htbp]
	\centering{\includegraphics[width=14cm]  {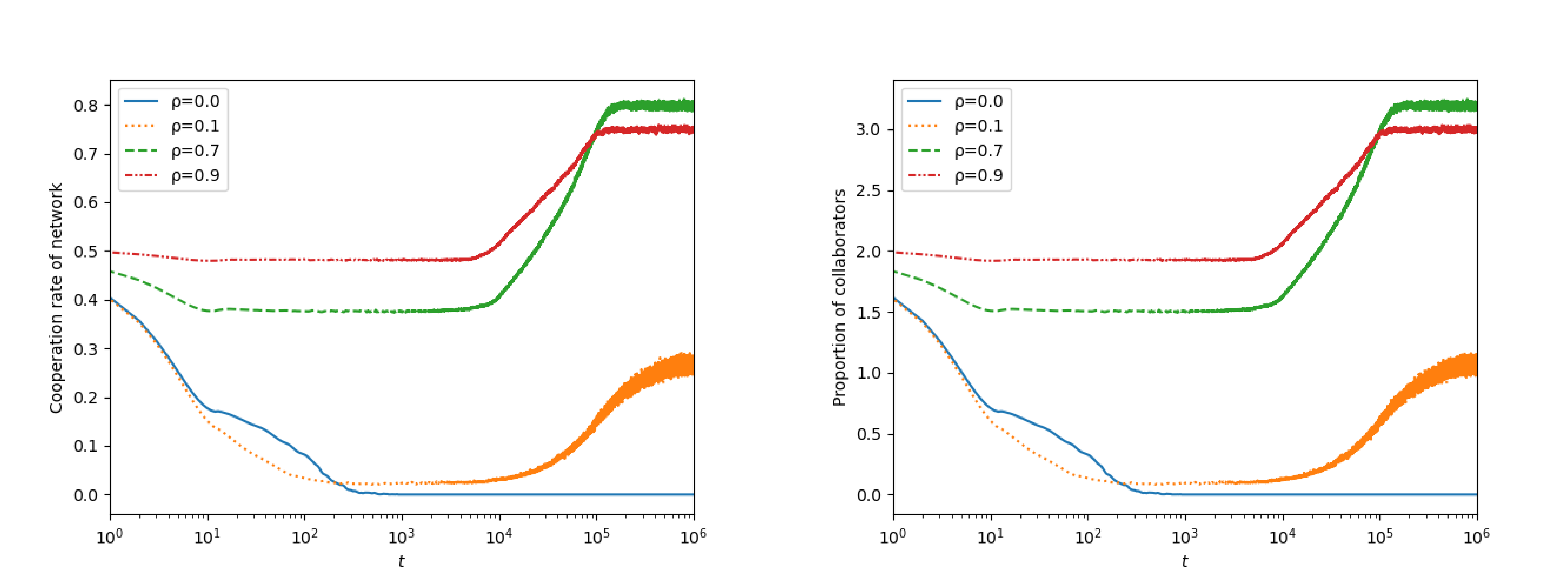}}
	\caption{\label{two_picture}Evolution of cooperation rate and reward under different $\rho$. The graph on the left is the cooperation rate, and the other is the reward. The trends of these two indicators are almost the same.}
\end{figure*}

\begin{figure}[h]
	\centering{\includegraphics[width=7cm]  {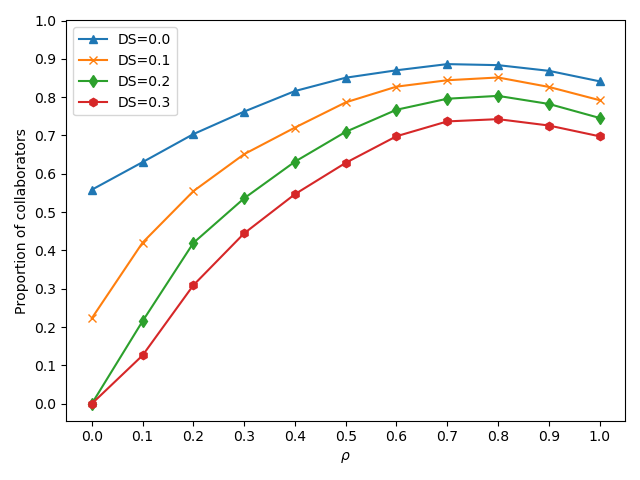}}
	\caption{\label{rate_of_collaborators}The cooperation level as a function of $\rho$ for different values of $DS$.}
\end{figure}

We can observe from Fig.\ref{rate_of_collaborators} that there is a trend of increasing first and then decreasing when $\rho$ increases. The cooperation rate is lowest when all agents on the network are traditional agents. Whereas when 70\%-80\% of agents are intelligent, the cooperation rate is the highest. Based on our experiments, we speculate that as the proportion of SARSA agents increases, more traditional agents attached to SARSA agents will tend to become cooperators. However, SARSA agents are intelligent individuals who will consider their benefits. If the surrounding individuals are all cooperators, they will tend to change from cooperators to betrayers to improve their rewards. Therefore, as the proportion of SARSA agents increases, the experiment shows a trend of the overall cooperation rate increasing and then decreasing. Longitudinally, as $DS$ increases, the value of the cooperation rate decreases continuously. That is consistent with the impact that $DS$ imparts in a conventional setting. From this Fig.\ref{rate_of_collaborators}, we learned that introducing agents that make autonomous decisions into the network increased the overall cooperation rate.

We plotted how the whole network cooperation rate changed during the convergence process to observe the network situation. The broken line diagram on the left side of Fig.\ref{two_picture} shows the change process of the cooperation rate. The line chart on the right shows the change in rewards. We can observe that the two trends are roughly the same. They all fall first and then rise. Due to the self-interest mechanism, the cooperation rate will show a downward trend. Then, with the continuous accumulation of experience with the SARSA algorithm, SARSA agents will tend to become partners in a group. It shows that SARSA agents can effectively enhance learning strategies while enhancing the cohesion of collaborators.

\section{Conclusion}\label{Conclusion}
In this paper,the SARSA algorithm, one of the classical reinforcement learning algorithms, is introduced into the research of evolutionary game-related problems. We apply the SARSA algorithm to the selection of learning objects in the game, which casts off the strong randomness and restrictions compared to the original rules. The participants will take ``income maximization'' as the goal, select the learning objects that can yield higher rewards according to the Q-table, and learn their strategies according to the probability. This approach can tremendously improve the cohesion between partners. Then, we let agents use the SARSA algorithm for autonomous decision-making. The process is similar to the above, except that the strategy chosen at this time is cooperation or betrayal. When SARSA agents have accumulated some experience, they tend to become partners, which makes the partners of traditional agents tend to rely on SARSA agents to form a cluster of partners. To sum up, the SARSA mechanism can effectively improve the cooperation rate in the network.

\acknowledgments This work is supported by the Natural Science Foundation of Fujian Province, China under Grant No. 2022J01117, Fujian Provincial Youth Education and Scientific Research Project under Grant No. JAT200007, and Fujian Key Laboratory of Data Science and Statistics (Minnan Normal University) under Grant No. 2020L0701.

\bibliographystyle{eplbib}
\bibliography{ref}

\end{document}